\documentclass{article} 
\usepackage{graphicx}
\usepackage{subfigure}
\usepackage{booktabs} 
\usepackage{algorithm}
\usepackage{algpseudocode}

\usepackage{hyperref}

\usepackage{amsmath}
\usepackage{amssymb}
\usepackage{mathtools}
\usepackage{amsthm}

\usepackage[utf8]{inputenc} 
\usepackage[T1]{fontenc}    
\usepackage{hyperref}       
\usepackage{url}            
\usepackage{booktabs}       
\usepackage{amsfonts}       
\usepackage{nicefrac}       
\usepackage{microtype}      
\usepackage{xcolor}         
\def\bbtheta{{\mbox{\boldmath $\theta$}}}

\def\bbSig{{\mbox{\boldmath $\Sigma$}}}
\def\bbphi{{\mbox{\boldmath $\phi$}}}

\def \nrf {D }
\def \noisevar {\sigma_{n}^2}

\def\bbXi{{\mbox{\boldmath $\Xi$}}}
\def\bbzeta{{\mbox{\boldmath $\zeta$}}}

\title{Weighted Ensembles for Active Learning with Adaptivity}

\author{
       Konstantinos D. Polyzos\thanks{The work of Konstantinos D. Polyzos was also supported by the Onassis Foundation Scholarship.} , Qin Lu, and Georgios B. Giannakis \\
       \textit{University of Minnesota} \\
       \texttt{\{polyz003, qlu, georgios\}@umn.edu}
}
  
\begin{document}
\maketitle

\begin{abstract}
Labeled data can be expensive to acquire in several application domains, including medical imaging, robotics, and computer vision. To efficiently train machine learning models under such high labeling costs, active learning (AL) judiciously selects the most informative data instances to label on-the-fly. This active sampling process can benefit from a statistical function model, that is typically captured by a  Gaussian process (GP). While most GP-based AL approaches rely on a single kernel function, the present contribution advocates an \emph{ensemble} of \emph{GP models} with weights adapted to the labeled data collected incrementally. Building on this novel EGP model, a suite of acquisition functions emerges based on the uncertainty and disagreement rules. An adaptively weighted \emph{ensemble} of EGP-based \emph{acquisition functions} is also introduced to further robustify performance. Extensive tests on synthetic and real datasets showcase the merits of the proposed EGP-based approaches with respect to the single GP-based AL alternatives.
\end{abstract}

\section{Introduction}
Identifying machine learning models usually relies on a sufficient number of labeled input-output data pairs, which may not be feasible due to labeling costs or privacy concerns in a number of application domains, ranging from healthcare~\cite{hoi2006batch}, computer vision~\cite{kapoor2007}, to robotics~\cite{taylor2021active}. To cope with sampling cost constraints, active learning (AL) selects a set of most informative data to label incrementally~\cite{settles2012,tong2001active}. This selection process can benefit from a statistical model for the learning function $f({\bf x})$ that maps each input feature vector ${\bf x}_\tau$ to the output/label $y_\tau$~\cite{cohn1996active}. Capable of learning nonlinear functions with {\it uncertainty quantification} in a {\it sample-efficient} fashion, Gaussian processes (GPs) are widely adopted to model the aforementioned $f$ in AL; see e.g.,~\cite{krause2008near, kapoor2007}. Given labeled samples 
$\{ \mathbf{x}_\tau,y_\tau\}_{\tau=1}^t$ collected in the set ${\cal L}_t$, GP modeling yields a function posterior probability density function (pdf) $p(f({\bf x})|{\cal L}_t)$. For regression, the latter is Gaussian with mean and variance available in closed form, while the model uncertainty captured by the variance is essential to guide the selection of subsequent instances to be queried.

The expressiveness of GP models depends on how well the chosen covariance (kernel) is adapted to the data at hand. Typically, GP-based AL approaches rely on a single kernel with {\rm preselected} form, which may exhibit limited expressiveness for AL, where labeled data are collected incrementally. On par with this online interactive operation, a key desideratum is a more expressive function space to improve adaptation to the labeled instances  on-the-fly. Besides expressiveness of  the function model, the AL performance is critically affected by the data acquisition rule. For regression, GP-AL typically leverages the function variance to select the next instance to be queried. Devising alternative acquisition rules for more expressive GP models is an open problem. Clearly, with multiple acquisition rules available, devising a strategy to combine and adapt them will be conducive to robust  performance across tasks.

To address the aforementioned challenges, our contributions here can be summarized in the following three aspects.
\begin{itemize}
    \item[c1)] Relative to GP-based AL that relies on a single GP with a {\it preselected} kernel, we introduce \emph{weighted ensembles} (E) of GPs for enhanced expressiveness with weights capturing the contributions of individual GP models, and adapting to labeled data collected on-the-fly. 
    \item[c2)] Utilizing the novel EGP model, we devise a suite of acquisition functions (AFs), including \emph{weighted ensembles} of AFs that further robustify performance.
    \item[c3)] Our thorough tests on synthetic and real data corroborate the impressive merits of GP and AF ensembles. 
\end{itemize}

\section{Related works}
This section outlines the context of the present work. 
\vspace{0.1cm}

\noindent{\bf Statistical models for AL.} The performance of model-based AL depends critically on the chosen statistical model. GPs are widely used because they come with uncertainty quantification and sample efficiency~\cite{krause2008near, kapoor2007}. However, most of the existing approaches rely on a preselected GP kernel with limited expressiveness, which may fall short in characterizing the incrementally collected data in AL. Broadening the scope of a single GP, a mixture of GPs is viable by training each GP component on a subset of labeled data ~\cite{zhao2020promoting}. This GP mixture model can account for multi-modalities in the function space, but needs to be refitted per iteration using nontrivial variational  methods. On the other hand, the proposed EGP model requires minimal refitting efforts, and trains each GP on the whole labeled set. Besides GP-based statistical models, Bayesian convolutional networks have been leveraged for image data~\cite{gal2017deep}.
\vspace{0.1cm}

\noindent{\bf Acquisition rules for AL.} 
The acquisition function (AF) also plays a performance - critical role in AL. Different AFs have been devised based on distinct selection criteria, learning tasks, as well as  whether or not a statistical model is capitalized on; see e.g., ~\cite{settles2012}. How to appropriately select the AF from the  available choices usually calls for domain expertise. It has been shown empirically that no acquisition rule excels in all tasks. This motivates a strategy that can combine and adapt candidate selection rules. In the closely related context of Bayesian optimization, multiple AFs have been combined to robustify performance across tasks~\cite{hoffman2011portfolio,shahriari2014entropy}. Such AF ensembles however, have not been explored for AL.
\vspace{0.1cm}

\noindent{\bf Kernel selection for GPs.} Adaptively choosing the form of the kernel from training data has been reported for conventional GP learning; see, e.g.,~\cite{teng2020scalable,duvenaud2013structure,kim2018scaling, malkomes2016bayesian}. These approaches usually operate in a batch offline mode, and rely on a large number of samples -- what discourages their use for AL, where data are not only acquired online, but are also scarce due to the potentially high labeling cost. Recently, an online scalable kernel selection approach has been introduced that leverages a set of GP experts in an online Bayesian model averaging fashion \cite{lu2020ensemble, lu2022incremental}. Although the notion of ensembles of GP experts has been explored in unsupervised learning \cite{karanikolas2021online}, graph-guided semi-supervised learning \cite{polyzos2021ensemble,polyzos2021graph,polyzos2021online}, reinforcement learning \cite{lu2021robust, lu2021gaussian, polyzos2021policy} and Bayesian optimization \cite{lu2022EGPBO}, to the best of our knowledge it has not yet been utilized in the AL context, which also entails the AF design. 

\section{Preliminaries}
This section outlines the motivation and preliminaries for the AL approach of interest. 

Typical learning approaches boil down to estimating the mapping $f(\cdot)$ that relates the $d\times 1$ input feature vector $\mathbf{x}_\tau$ to the output $y_\tau$ (that is either a real number in regression or it belongs to a finite alphabet in classification) as $\mathbf{x}_\tau \rightarrow f(\mathbf{x}_\tau)\rightarrow y_\tau$. This estimation task relies on a sufficient number of labeled training samples $\{ \mathbf{x}_\tau, y_\tau\}_{\tau=1}^T$. In several applications however, the input can be readily obtained whereas the corresponding label can be expensive to acquire due to sampling costs or privacy concerns. In healthcare for instance,  many labels describing the medical condition of patients may not be revealed to preserve confidentiality. Faced with this challenge, one can resort to the AL paradigm, which judiciously and proactively selects the most {\it informative} instances to label so that the sought mapping can be inferred in a {\it sample-efficient} manner.

AL starts with a small-size set $\mathcal{L}_0 := \{(\mathbf{x}_\tau, y_\tau)\}_{\tau=-L_0+1}^{0}$ of labeled samples,\footnote{The negative instance index here is used for notational brevity as more labeled data are included next.} and a larger collection of unlabeled features $\mathcal{U}_0 := \{\mathbf{x}_\tau^u \}_{\tau=1}^{U_0}$ ($L_0\ll U_0$). Given corresponding sets $\mathcal{L}_t$ and $\mathcal{U}_t$ up to index $t>0$, model-based AL entails a statistical function model, namely the pdf $p(f({\bf x})|{\cal L}_t)$. The latter is utilized by the so-termed  {\it acquisition function} (AF) $\alpha (\cdot)$ to select the instance $\mathbf{x}_{t+1}$ from the corresponding unlabeled set $\mathcal{U}_t$, as~\cite{cohn1996active}
\begin{align}
{\bf x}_{t+1} =  \underset{\mathbf{x}\in\mathcal{U}_t}{\arg\max} \; \; \alpha(\mathbf{x}; {\cal L}_t)  \;.\label{eq:AF_1}
\end{align}
Intuitively, $\alpha$ is chosen to guide exploration of the space $f(\cdot)$ belongs to, which hinges on quantifying the uncertainty of the belief model $p(f({\bf x})|{\cal L}_t)$.
Upon querying an {\it oracle} for the associated label $y_{t+1}$, the labeled set is then augmented with the new pair and the feature vector is removed from the unlabeled set, that is ${\cal L}_{t+1}:= {\cal L}_{t} \cup \{(\mathbf{x}_{t+1}, y_{t+1} ) \}$ and ${\cal U}_{t+1}:= {\cal U}_{t} \setminus \{\mathbf{x}_{t+1}\}$. Apparently, the two critical choices are the model for $f$, and the AF design for $\alpha$. Focusing on the regression task, we will outline the GP-based model for $f$, and the associated acquisition rules next.


\subsection{GP-based active learning }
GPs estimate a nonparametric function model in a sample-efficient manner, while also offering quantification of the associated model uncertainty \cite{Rasmussen2006gaussian}. Sample efficiency justifies their wide adoption in AL. The GP model postulates $f$ as being randomly drawn from a GP prior; that is $f\sim \mathcal{GP}(0, \kappa(\mathbf{x},\mathbf{x}'))$ with $\kappa(\mathbf{x},\mathbf{x}')$ being a positive-definite kernel function that measures the pairwise similarity between two distinct inputs $\mathbf{x}$ and $\mathbf{x}'$. With $^\top$ denoting transposition, this means that the random vector $\mathbf{f}_t := [f(\mathbf{x}_1) \ldots  f(\mathbf{x}_t)]^\top$ consisting of the function evaluations at instances $\mathbf{X}_t := \left[\mathbf{x}_1  \ldots \mathbf{x}_t\right]^\top$ is Gaussian distributed as \cite{Rasmussen2006gaussian}
\begin{equation}
p(\mathbf{f}_t| \mathbf{X}_t) = \mathcal{N} (\mathbf{f}_t ; {\bf 0}_t, {\bf K}_t )\  \  \ \forall t \nonumber
\label{eq:gp_prior}
\end{equation}
where $\mathbf{K}_t$ denotes the $t \times t$ covariance matrix whose $(i,j)$ entry  is $[{\bf K}_t]_{i,j} = {\rm cov} (f(\mathbf{x}_i), f(\mathbf{x}_j)):=\kappa(\mathbf{x}_i, \mathbf{x}_j)$.

Next, the (possibly noisy) output data $\mathbf{y}_t := [y_1 \cdots y_t]^\top$ are related to the function evaluations $\mathbf{f}_t$ through the likelihood $p(\mathbf{y}_t|\mathbf{f}_t ; \mathbf{X}_t)$ which is assumed to be factored as $p(\mathbf{y}_t|\mathbf{f}_t; \mathbf{X}_t) = \prod_{\tau=1}^t p(y_\tau|f(\mathbf{x}_\tau))$ with $p(y_\tau|f(\mathbf{x}_\tau))$ denoting the per-datum known likelihood factor. The latter holds in the regression setup, where $p(y_\tau|f(\mathbf{x}_\tau)) = \mathcal{N}(y_\tau;f(\mathbf{x}_t),\sigma_n^2)$, so long as $y_\tau$ is expressed as $y_\tau = f_\tau + n_\tau$ with $n_\tau \sim \mathcal{N}(n_\tau;0,\sigma_n^2)$ being white Gaussian noise uncorrelated across instances. For the regression task, predicting the function value of a generic input $\mathbf{x}$ is carried out after writing the joint pdf of $\mathbf{y}_t$ and $f({\bf x})$ as
\begin{align}
&\begin{bmatrix}     \mathbf{y}_t \\  f({\bf x}) \end{bmatrix}\!\sim\!  \mathcal{N}\!\left(\!
\mathbf{0}_{t+1},\!\!
\begin{bmatrix} 
\mathbf{K}_t\!+\!\sigma_n^2 \mathbf{I}_t&  \mathbf{k}_t ({\bf x})  \\
\mathbf{k}_{t}^\top ({\bf x})  & \kappa(\mathbf{x},\mathbf{x})\!+\!\sigma_n^2
\end{bmatrix}
\!   \right)   \nonumber\label{eq:joint_pdf}
\end{align}
where $\mathbf{k}_t ({\bf x}) := [\kappa(\mathbf{x}_1, \mathbf{x}), \ldots, \kappa(\mathbf{x}_t,  \mathbf{x})]^\top$.

With the joint pdf at hand, the posterior pdf of $f({\bf x})$ is 
\begin{align}
p(f({\bf x})|\mathbf{y}_t;\mathbf{X}_{t}) = \mathcal{N}(f({\bf x}); \mu_t ({\bf x}), \sigma_{t}^2 ({\bf x}))
\end{align} 
where 
\begin{subequations}
\begin{align}	
\mu_t ({\bf x}) & = \mathbf{k}_t^{\top} ({\bf x}) (\mathbf{K}_t + \sigma_{n}^2
 \mathbf{I}_t)^{-1} \mathbf{y}_t \label{eq:mean}\\
\sigma_{t}^2 ({\bf x})& = \!\kappa(\mathbf{x},\mathbf{x})\! -\! \mathbf{k}_t^{\top} ({\bf x}) (\mathbf{K}_t\! +\! \sigma_{n}^2 \mathbf{I}_t)^{-1} \mathbf{k}_t ({\bf x}) .\label{eq:variance}
\end{align}\label{eq:plain_gpp}
\end{subequations}
Note that the mean in \eqref{eq:mean} is a point prediction of $f({\bf x})$, while the variance in \eqref{eq:variance} quantifies the associated uncertainty. In the AL context, this uncertainty is used by the acquisition function that selects the next query instance as
\begin{align}
{\bf x}_{t+1} =  \underset{\mathbf{x}\in\mathcal{U}_t}{\arg\max} \; \; \sigma_{t}^2 ({\bf x})\;. \label{eq: singleGPvaracq}
\end{align}
It is worth mentioning that for a Gaussian pdf, \eqref{eq: singleGPvaracq} is tantamount to maximizing the entropy~\cite{mackay1992information}.

The posterior mean and variance in~\eqref{eq:plain_gpp} rely on all $t$ instances in $\mathbf{X}_t$ to form $\mathbf{K}_t$, and the associated complexity for its inversion is thus  $\mathcal{O}(t^3)$. Although this complexity can be affordable in AL where $t$ is small, it can be further reduced. In addition, $\mu_t ({\bf x})$ and $\sigma_{t}^2 ({\bf x})$ require direct access to $\{ \mathbf{x}_\tau\}_{\tau = 1}^t$, which may be discouraged due to privacy concerns as in e.g medical records and financial statements. Further, GP-based AL relies on a  preselected kernel function, which may exhibit limited expressiveness. These limitations can be ameliorated through our novel ensemble approach that leverages also random spectral features, as delineated next.


\begin{algorithm}[t]
\caption{RF-based EGP-AL.} 
\label{Alg: EGP-AL}
\begin{algorithmic}[1]
\State \textbf{Initialization}: $\mathcal{U}_0$, $\mathcal{L}_0$, $\mathcal{K}$; 
\vspace{0.05cm}
\For{$t = 0, \ldots, T$}
\State Obtain $p(f({\bf x})|{\cal L}_t)$ via $\bbXi_t$;
\State Select ${\bf x}_{t+1}$ based on one from~\eqref{eq:w_var}-~\eqref{eq:QBC},~\eqref{eq:var-GPM}-\eqref{eq:mix.ent.};
\State Query the oracle to obtain $y_{t+1}$;
\State ${\cal L}_{t+1}\!\!:=\! {\cal L}_{t} \cup \{(\mathbf{x}_{t+1}, y_{t+1} ) \}$, ${\cal U}_{t+1}\!\!:=\! {\cal U}_{t} \setminus \{\mathbf{x}_{t+1}\}$;
\EndFor
\end{algorithmic}
\end{algorithm}

\section{Ensemble GPs for AL}

The chosen function model affects critically the performance of AL approaches. Unlike most existing works that rely on a single GP with a {\it preselected} kernel, we advocate an ensemble (E) of $M$ GPs to enhance expressiveness. Each GP has a distinct kernel function selected from a given dictionary $\mathcal{K}:=\{\kappa^1, \ldots, \kappa^M\}$, that is formed using kernels of different types and with different hyperparameters. Specifically, each GP $m\in \mathcal{M}:=\{1,\ldots,M\}$ places a unique prior on $f$ as $f|m \sim \mathcal{GP}(0, \kappa^m (\mathbf{x}, \mathbf{x}'))$. The EGP prior of $f(\mathbf{x})$ is then a weighted ensemble of the individual GP priors as
\begin{align}
	f(\mathbf{x})\sim \sum_{m=1}^M w^m_0 {\cal GP}(0,\kappa^m (\mathbf{x},\mathbf{x}')),  \;\;\;\; \sum_{m=1}^M w^m_0 =1  \label{eq:EGP_prior}
\end{align}\\
where $w^m_0:={\rm Pr} (i=m)$ is the prior probability that measures the contribution of GP model $m$. With labeled data collected on-the-fly, the sum-product rule allows one to express the EGP-based function posterior pdf as
\begin{align}
	{p}(f(\mathbf{x})|\mathcal{L}_{t}) \! &=\! \sum_{m = 1}^M\! {\rm Pr}(i\!=\! m|\mathcal{L}_{t}) {p}(f(\mathbf{x})|  i\!=\! m,\mathcal{L}_{t} ) \label{eq:EGP_post}
	\end{align}
which is a mixture of posterior GPs with weights $w_t^m:={\rm Pr} (i=m|{\cal L}_t)$ that signify the significance of the GP experts. These weights thus enable online model adaptation.  


To efficiently update this EGP function model across $t$, we will leverage a parametric function approximant, formed by the so-termed random features (RFs), as outlined next.

\subsection{RF-based approximation per GP}

The RF-based approximation relies on a shift-invariant kernel $\bar{\kappa}(\mathbf{x}, \mathbf{x}') = \bar{\kappa}(\mathbf{x}-\mathbf{x}')$ satisfying $\kappa = \sigma_{\theta}^2\bar{\kappa}$, which can be expressed as the inverse Fourier transform of a spectral density $\pi_{\bar{\kappa}}(\bbzeta)$ as ~\cite{rudin1964principles}
\begin{align}
 \bar{\kappa}(\mathbf{x}-\mathbf{x}') = \int \pi_{\bar{\kappa}} (\bbzeta) e^{j\bbzeta^\top (\mathbf{x} - \mathbf{x}')} d\bbzeta = \mathbb{E}_{\pi_{\bar{\kappa}}} \left[e^{j\bbzeta^\top (\mathbf{x} - \mathbf{x}')} \right] \nonumber
\end{align}
where $\int \pi_{\bar{\kappa}}(\bbzeta)d \bbzeta=1$, allowing $\pi_{\bar{\kappa}}$ to be deemed as a pdf. Since $\bar{\kappa}$ is real, the last expectation is equal to $\mathbb{E}_{\pi_{\bar{\kappa}}} \left[\cos(\bbzeta^\top (\mathbf{x} - \mathbf{x}'))\right]$; and after drawing a sufficient number $\nrf$ of  independent and identically distributed (i.i.d.) samples $\{\bbzeta_j \}_{j = 1}^{\nrf}$ from $\pi_{\bar{\kappa}} (\bbzeta)$, kernel $\bar{\kappa}$ is approximated by
\begin{align}
	\check{\bar{\kappa}} (\mathbf{x}, \mathbf{x}'):= \frac{1}{\nrf} \sum_{j = 1}^{\nrf} \cos  \left(\bbzeta_j^\top (\mathbf{x} - \mathbf{x}') \right) \;. \label{kern_est}
\end{align}
Defining the $2\nrf\!\times\! 1$ RF vector as~\cite{quia2010sparse}
\vspace*{-0.15cm} 
\begin{align}
	&\bbphi_{\bbzeta} (\mathbf{x})  \label{eq:phi_x}\\
	&:=\! \frac{1}{\sqrt{\nrf}}\!\left[\sin(\bbzeta_1^\top \mathbf{x}), \cos(\bbzeta_1^\top \mathbf{x}), \ldots, \sin(\bbzeta_{\nrf}^\top \mathbf{x}), \cos(\bbzeta_{\nrf}^\top \mathbf{x})\right]^{\top} \nonumber
\end{align}
the sample average~\eqref{kern_est} can be re-written as $\check{\bar{\kappa}}(\mathbf{x}, \mathbf{x}') = \bbphi_{\bbzeta}^{\top} (\mathbf{x})\bbphi_{\bbzeta}(\mathbf{x}')$. This allows the \emph{parametric linear} function
\begin{align}
	{\check f} (\mathbf{x}) =  \bbphi_{\bbzeta}^\top (\mathbf{x}) \bbtheta, \quad \bbtheta\sim \mathcal{N}(\bbtheta; \mathbf{0}_{2\nrf}, \sigma_\theta^2\mathbf{I}_{2\nrf})\; \label{eq:f_check}
\end{align}
to have an approximate GP prior. Such an RF-based parametric function readily yields an efficient model update by propagating the posterior pdf $p(\bbtheta|\mathbf{y}_{t};\mathbf{X}_t) = \mathcal{N}(\bbtheta; \hat{\bbtheta}_t, \bbSig_t)$ per slot $t$ in a recursive Bayes fashion.  It is also worth stressing that the model learning step does not require direct access to $\mathbf{x}_t$, but relies only on the RF vector $\boldsymbol{\phi}_{\bbzeta}(\mathbf{x}_{t})$, which can be viewed as an encrypted version of $\mathbf{x}_{t}$. This may be appealing if $\mathbf{x}_{t}$, which may e.g., comprise private medical data, cannot be revealed during model learning.

Having outlined the RF-based approximation per GP, we can proceed with updating our RF-based EGP function model, as labeled instances become available incrementally.
 


\subsection{EGP parametric model updates}
When kernels in the dictionary are shift-invariant, the RF vector $\bbphi_{\bbzeta}^m(\mathbf{x})$ per GP $m$ can be formed via~\eqref{eq:phi_x} by first drawing i.i.d. random vectors $\{\bbzeta_j^m\}_{j=1}^{\nrf}$ from $\pi_{\bar{\kappa}}^m (\bbzeta)$, which is the spectral density of the standardized kernel $\bar{\kappa}^m$. Let  $\sigma_{\theta^m}^2$ be the kernel magnitude, so that $\kappa^m = \sigma_{\theta^m}^2\bar{\kappa}^m$. The generative model for the sought function and the noisy output $y$  per GP $m$ are characterized by the $2D\times 1$ vector $\bbtheta^m$ as
	\begin{align}
		{p}(\bbtheta^m) &= \mathcal{N} (\bbtheta^m; \mathbf{0}_{2\nrf}, \sigma_{\theta^m}^2\mathbf{I}_{2\nrf})  \nonumber\\
		p(f(\mathbf{x}_\tau)|i=m,\bbtheta^m )& = \delta (f(\mathbf{x}_\tau)-\bbphi_{\bbzeta}^{m\top} (\mathbf{x}_\tau)\bbtheta^m)\nonumber\\
		p(y_\tau|\bbtheta^m, \mathbf{x}_\tau) &=
		\mathcal{N}(y_\tau; \bbphi_{\bbzeta}^{m\top} (\mathbf{x}_\tau)\bbtheta^m, \sigma_n^2) \;.\label{eq:LF}
	\end{align}
This parametric form allows one to capture the function posterior pdf per GP $m$ via $p(\bbtheta^m|\mathcal{L}_t) = \mathcal{N}(\bbtheta^m; \hat{\bbtheta}_t^m, \bbSig_t^m)$, which together with the weight $w_t^m$, approximates the EGP function posterior~\eqref{eq:EGP_post} via
\begin{align}
p(\check{f}({\bf x})|{\cal L}_t)=\sum_{m=1}^M w_t^m {\cal N}(\check{f}({\bf x}); \check{\mu}_t^m ({\bf x}), (\check{\sigma}_t^m ({\bf x}))^2 )\label{eq:RF_func_post}
\end{align}
with
\begin{subequations}
\begin{align}
\check{\mu}_t^m ({\bf x})& =  \bbphi^{m\top}_{\bbzeta} (\mathbf{x})\hat{\bbtheta}_t^{m} \label{eq:mu_check}\\
(\check{\sigma}_t^m ({\bf x}))^2 & =  \bbphi^{m\top}_{\bbzeta}  (\mathbf{x}) \bbSig^m_{t} \bbphi^m_{\bbzeta} (\mathbf{x})\;.\label{eq:sigma_check}
\end{align}
\end{subequations}
Next, we will see how RF-based EGP propagates the  function model by updating across $t$ the parameter set
\begin{align}
 \bbXi_t:=\{w_t^m, \hat{\bbtheta}_t^m, \bbSig_t^m, m\in\mathcal{M}\}   \;.\label{eq:xi_t}
\end{align}

Upon acquiring the newly labeled pair $\{\mathbf{x}_{t+1},y_{t+1}\}$, the updated weight $w_{t+1}^m := {\rm Pr}(i=m|\mathcal{L}_{t+1})$ can be obtained per GP $m$ via Bayes' rule as
\begin{align}
	w_{t+1}^m &
	= \frac{{\rm Pr}(i=m|\mathcal{L}_{t}) {p}(y_{t+1}|\mathbf{x}_{t+1},  i=m,\mathcal{L}_{t})}{{p}(y_{t+1}|\mathbf{x}_{t+1}, \mathcal{L}_{t})}  . \nonumber
\end{align}
Since the per-model predictive likelihood is given by
\begin{align}
	{p}(y_{t+1}|i\!=\!m,\mathcal{L}_{t}, \mathbf{x}_{t+1})    \!\! &=\!\! \int\!\! p(y_{t+1}| \bbtheta^m\! ,\mathbf{x}_{t+1}) 
	p(\bbtheta^m|\mathcal{L}_t) d \bbtheta^m \nonumber \\
	& = \mathcal{N}(y_{t+1};\hat{y}_{t+1|t}^{m},(\sigma_{t+1|t}^{m})^2)\nonumber
\end{align}\vspace*{-0.5cm}\\
with 
\begin{subequations}
	\label{eq:pred_y}
	\begin{align}
		\hat{y}_{t+1|t}^{m} &=  \bbphi^{m\top}_{\bbzeta} (\mathbf{x}_{t+1})\hat{\bbtheta}_t^{m} \nonumber\\	
		(\sigma_{t+1|t}^{m})^2 & = \bbphi^{m\top}_{\bbzeta}  (\mathbf{x}_{t+1}) \bbSig^m_{t} \bbphi^m_{\bbzeta} (\mathbf{x}_{t+1})+\noisevar\nonumber
	\end{align}
\end{subequations}
the updated weight can thus be expressed as
\begin{align}
	w_{t+1}^m  & = \frac{w_t^m \mathcal{N}\left(y_{t+1};  \hat{y}_{t+1|t}^{m}, (\sigma_{t+1|t}^{m})^2 \right)}{\sum_{m' = 1}^M w_t^{m'} \mathcal{N}\left(y_{t+1};  \!\hat{y}_{t+1|t}^{m'}, (\sigma_{t+1|t}^{m'})^2 \right)}  \;. \label{eq:w_update}
\end{align}
Bayes' rule further allows updating the posterior of $\bbtheta^m$ as 
\begin{align}
{p}(\bbtheta^m |\mathcal{L}_{t+1}) & =  	\frac{{p}(\bbtheta^m |\mathcal{L}_{t}\!) {p}(y_{t+1}|\bbtheta^m,\mathbf{x}_{t+1})}
{{p}(y_{t+1}|\mathbf{x}_{t+1},  i=m,\mathcal{L}_{t})}\nonumber\\
& = \mathcal{N}(\bbtheta^m; \hat{\bbtheta}_{t+1}^m, \bbSig^m_{t+1}) \label{eq:posterior_update_1}
\end{align}
where the mean $ \hat{\bbtheta}_{t+1}^m$ and covariance matrix $\bbSig^m_{t+1}$ are
\begin{subequations} 
	\begin{align}
		\hspace*{-0.25cm}\hat{\bbtheta}_{t+1}^m &\!=\! \hat{\bbtheta}_{t}^m \!+\!  (\!\sigma_{t+1|t}^{m})^{-2}\bbSig^m_{t} \bbphi^m_{\bbzeta}(\mathbf{x}_{t+1})(y_{t+1} \!-\! \hat{y}_{t+1|t}^{m}) \nonumber \\
	\hspace*{-0.25cm}	\bbSig_{t+1}^m &\!\!=\! \bbSig_{t}^m\! \!-\!  (\!\sigma_{t+1|t}^{m})^{-2}\bbSig^m_{t} \bbphi^m_{\bbzeta}\!(\mathbf{x}_{t+1})  \bbphi^{m\top}_{\bbzeta}\!\!(\mathbf{x}_{t+1})\bbSig^m_{t}\!. \nonumber
	\end{align}
\end{subequations}

\subsection{Acquisition rules for EGP-based AL}
Using the EGP posterior in~\eqref{eq:RF_func_post}, we are ready to devise AFs that select the next query point based on different rules.

\subsubsection{Weighted variance}
Motivated by~\eqref{eq: singleGPvaracq}, the first AF relies on the uncertainty expressed by the variance. With GP expert $m$ forming the function posterior with variance $(\check{\sigma}_t^m ({\bf x}))^2$, a weighted combination over all the $M$ experts yields the AF
\begin{align}
\alpha^{\rm wVar} ({\bf x}; {\cal L}_t) := \sum_{m=1}^M w_t^m (\check{\sigma}_t^m ({\bf x}))^2 \;. \label{eq:w_var}
\end{align}

\subsubsection{Weighted entropy}
Relying alternatively on the entropy as the uncertainty measure, one can take a weighted sum of the entropy values given by the $M$ GP experts, yielding
\begin{align}
\alpha^{\rm wEnt} ({\bf x}; {\cal L}_t) :=\frac{1}{2} \sum_{m=1}^M w_t^m \ln(2\pi(\check{\sigma}_t^m ({\bf x}))^2)\;. \label{eq:w_entropy}
\end{align}

\subsubsection{Query-by-Committee (QBC)}
Besides capturing uncertainty by variance or entropy, an alternative disagreement-based AF -- QBC, has been reported for classification~\cite{seung1992query}, and regression using neural networks~\cite{krogh1994neural}. With the $M$ GP experts forming a committee, 
the novel EGP-based QBC rule is (cf.~\eqref{eq:mu_check})
\begin{align}
\alpha^{\rm QBC} ({\bf x}; {\cal L}_t) := \sum_{m=1}^M w_t^m (\check{\mu}_t^m ({\bf x})-\bar{\mu}_t ({\bf x}))^2   \label{eq:QBC}
\end{align}
where $ \bar{\mu}_t ({\bf x})$ is the consensus of the committee given by
\begin{align}
 \bar{\mu}_t ({\bf x}) = \sum_{m=1}^M w_t^m \check{\mu}_t^m ({\bf x}) \;. \label{eq:QBC_mean}
\end{align}
Unlike previous QBC approaches that have equal weights per committee member, the weights $w_t^m$ in ~\eqref{eq:QBC} and \eqref{eq:QBC_mean} are generally different across $m$.

\subsubsection{Variance of GP mixtures}
Rather than directly summing per-GP weighted variances in~\eqref{eq:w_var}, one can alternatively obtain the variance based on the GP mixture of the function posterior (cf.~\eqref{eq:RF_func_post}) as
\begin{align}
 &\alpha^{\rm GPM-Var} ({\bf x}; {\cal L}_t) \nonumber\\ 
 &:= \sum_{m=1}^M w_t^m ((\check{\sigma}_t^m ({\bf x}))^2 + (\check{\mu}_t^m ({\bf x})-\bar{\mu}_t ({\bf x}))^2   \label{eq:var-GPM}
\end{align}
which, interestingly, is the sum of~\eqref{eq:w_var} and~\eqref{eq:QBC}.

\subsubsection{Entropy of GP mixtures}
The last AF is given by the entropy of the GP mixture in~\eqref{eq:RF_func_post}, which unfortunately, has no analytic expression. Aiming at a tractable form, we will resort to its analytic lower bound~\cite{huber2008entropy}, which can be expressed as
\begin{align}
&-\!\!\sum_{m=1}^M\! w_t^m\!\! \int\!\! {\cal N}(\check{f}({\bf x});\check{\mu}_t^m ({\bf x}),\!(\check{\sigma}_t^m ({\bf x}))^2)\log p(\check{f}({\bf x})|\!{\cal L}_t) d\check{f}({\bf x})\nonumber\\
&\overset{(a)}{\geq} -\sum_{m=1}^M w_t^m \log \Big(\int {\cal N}(\check{f}({\bf x});\check{\mu}_t^m ({\bf x}),(\check{\sigma}_t^m ({\bf x}))^2)  \nonumber\\
&\qquad\qquad\qquad\qquad\quad \times p(\check{f}({\bf x})|{\cal L}_t) d\check{f}({\bf x})\Big) \nonumber
\end{align}
where $(a)$ holds due to Jensen's inequality. Upon obtaining the analytic expression for the term inside the logarithm, the last AF is then given by
\begin{align}
 \alpha^{\rm GPM-Ent} ({\bf x}; {\cal L}_t):=   -\sum_{m=1}^M w_t^m\log \left(\sum_{m'=1}^M w_t^{m'}z^{m,m'}_t\right)  \label{eq:mix.ent.}
\end{align}
with $z^{m,m'}$ accounting for the interaction of any two GP models as
\begin{align}
  z^{m,m'}_t&:=\int {\cal N}(\check{f}({\bf x});\check{\mu}_t^m ({\bf x}),(\check{\sigma}_t^m ({\bf x}))^2) \nonumber\\
  &\quad \times{\cal N}(\check{f}({\bf x});\check{\mu}_t^{m'} ({\bf x}),(\check{\sigma}_t^{m'} ({\bf x}))^2) d\check{f}({\bf x})\nonumber\\
  & ={\cal N}(\check{\mu}_t^m ({\bf x}); \check{\mu}_t^{m'} ({\bf x}),(\check{\sigma}_t^{m} ({\bf x}))^2+(\check{\sigma}_t^{m'} ({\bf x}))^2)\nonumber\;.
\end{align}
Based on our novel EGP-based AFs, implementation of the proposed EGP-AL approach is summarized in Alg.~1.

\begin{algorithm}[t]
\caption{EGP-MultiAFs for AL.} 
\label{Alg: EGP_multiAcq}
\begin{algorithmic}[1]
\State \textbf{Initialization:} $\mathcal{L}_0$, $\mathcal{U}_0$, $\mathcal{V}$, $\mathcal{K}$;
\State $\boldsymbol{\omega}_0 =\frac{1}{K} [1,\ldots, 1]^\top$;
\vspace{0.1cm}
\For{$t = 0, 1, \ldots, T$}
\vspace{0.1cm}
\State Obtain EGP $\bbXi_t$ based on ${\cal L}_t$ using (15)-(16);
\For{$k = 1, \ldots, K$}
\vspace{0.1cm}
\State Obtain instance $\tilde{\mathbf{x}}_{t+1}^{k} \in \mathcal{U}_{t}$ by ~\eqref{eq:AF_k}; 
\vspace{0.1cm}
\State Obtain pseudo-label $\tilde{y}_{t+1}^k$ using $\bbXi_t$ via~\eqref{eq:pseudo_k};
\vspace{0.1cm}
\State Using pseudo pair $\{\tilde{{\bf x}}_{t+1}^k, \tilde{y}_{t+1}^k\}$ obtain $\tilde{\bbXi}_{t+1}^k$;
\vspace{0.1cm}
\State Obtain error $\epsilon_{t+1}^{v,k}$ on $\mathcal{V}$ via~\eqref{eq:error_k};
\vspace{0.1cm}
\EndFor
\vspace{0.1cm}
\State Update per AF weight using~\eqref{eq:AF_weight}; 
\vspace{0.05cm}
\State Obtain $\mathbf{x}_{t+1} \in \mathcal{U}_{t}$ by~\eqref{eq: EGPmultiacq};
\vspace{0.1cm}
\State Query the oracle to obtain true label $y_{t+1}$;
\vspace{0.1cm}
\State $\mathcal{L}_{t+1} = \mathcal{L}_{t} \cup(\mathbf{x}_{t+1},y_{t+1})$;
\vspace{0.1cm}
\State $\mathcal{U}_{t+1} = \mathcal{U}_{t} \setminus \{\mathbf{x}_{t+1}\}$; 
\EndFor
\end{algorithmic}
\end{algorithm}

\section{Ensemble of EGP-based AFs}
So far, we have introduced a novel EGP-based function model along with several choices for the AF. In the context of Bayesian optimization though, it is known that no single AF excels at all tasks~\cite{hoffman2011portfolio}. Hence, combining candidate AFs can intuitively offer robustness and improved performance. To this end, we will rely on a validation set ${\cal V}:=\{(\mathbf{x}_\tau^v,y_\tau^v)\}_{\tau=1}^V$ to evaluate the performance of different AFs. Similar to EGP, each of the $K$ candidate AFs will come with a weight (probability) $\omega_t^k\in [0,1]$ to capture its contribution per slot $t$, such that $\sum_{k=1}^K \omega_t^k = 1$.

\begin{figure*}[t]
\centering
	\includegraphics[width=0.98\linewidth]{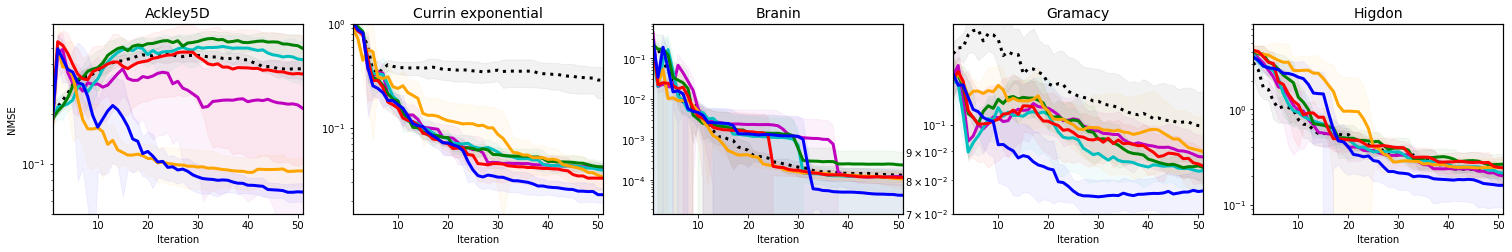}
	\includegraphics[width=0.99\linewidth]{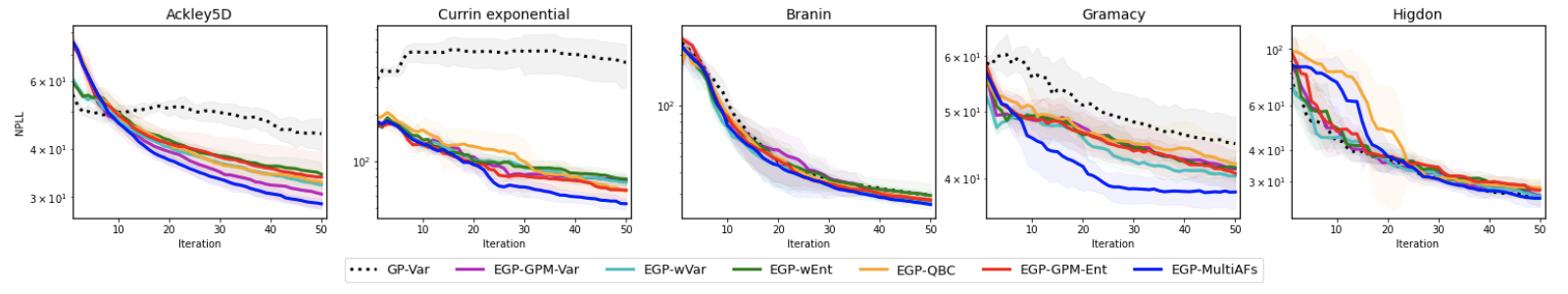}
	\vspace*{-0.1cm}
	\caption{NMSE and NPLL versus iterations for Ackley5D, Currin exponential, Branin, Gramacy and Higdon functions } \label{fig:NMSE_synFunc}
\end{figure*}

Upon identifying the RF-based EGP set $\bbXi_t$ in~\eqref{eq:xi_t} using the labeled set ${\cal L}_t$ at slot $t$, each AF $k$ selects its query point $\tilde{{\bf x}}_{t+1}^k$ at slot $t+1$ by optimizing the associated criterion as
\begin{align}
  \tilde{{\bf x}}_{t+1}^k = \underset{\mathbf{x}\in\mathcal{U}_t}{\arg\max} \; \; \alpha^k (\mathbf{x}; {\cal L}_t) \;. \label{eq:AF_k}
\end{align}
Upon obtaining $\tilde{{\bf x}}_{t+1}^k$, AF $k$ constructs a {\it `pseudo label'} $\tilde{y}_{t+1}^k$ using the EGP parameters in $\bbXi_t$, as
\begin{align}
  \tilde{y}_{t+1}^k = \sum_{m=1}^M w_t^m  \bbphi^{m\top}_{\bbzeta} (\tilde{{\bf x}}_{t+1}^k)\hat{\bbtheta}_t^{m}\;. \label{eq:pseudo_k}
\end{align}
This pseudo pair $\{\tilde{{\bf x}}_{t+1}^k, \tilde{y}_{t+1}^k\}$ allows one to leverage~\eqref{eq:w_update}--~\eqref{eq:posterior_update_1}  to find the updated EGP parameter vector as
\begin{align}
\tilde{\bbXi}_{t+1}^k=\{\tilde{w}_{t+1}^{m,k}, \tilde{\bbtheta}_{t+1}^{m,k}, \tilde{\bbSig}_{t+1}^{m,k}, m\in\mathcal{M}\}    \label{eq:EGP_parameter_k}
\end{align}
based on which the loss per AF can be evaluated. 

To find this loss, AF $k$ capitalizes on $\tilde{\bbXi}_{t+1}^k$ in order to obtain the prediction error at the validation set
\begin{align}
    \epsilon_{t+1}^{v,k}= V^ {-1} \sum_{\tau=1}^V (y_\tau^v -\hat{y}_{\tau|t+1}^{v,k})^2 \label{eq:error_k}
\end{align}
where the predicted label per validation sample $\tau$ is 
\begin{align}
\hat{y}_{\tau|t+1}^{v,k} = \sum_{m=1}^M \tilde{w}_{t+1}^{m,k} \bbphi^{m\top}_{\bbzeta} ({\bf x}_{\tau}^v)\tilde{\bbtheta}_{t+1}^{m,k} \;.
\end{align}
Having available the prediction error over the validation set per AF $k$, the associated weight can then be updated as
\begin{align}
    \omega_{t+1}^k = \frac{\omega_{t}^k \exp(-\eta \epsilon_{t+1}^{v,k})}{\sum_{k'=1}^K \omega_{t}^{k'} \exp(-\eta \epsilon_{t+1}^{v,k'})}\label{eq:AF_weight}
\end{align}
where $\eta$ denotes the learning rate.
Here, the weight update rule is similar to that in EGP (cf.~\eqref{eq:w_update}), and belongs to the exponential weight update in online learning with expert advice; see e.g.,~\cite{cesa2006prediction}.

Given the updated weights, the next query point is identified by maximizing the weighted ensemble of AFs, that is
\begin{align}
{\bf x}_{t+1} = \underset{\mathbf{x}\in\mathcal{U}_t}{\arg\max}  \sum_{k=1}^K \omega_{t+1}^k \alpha^k (\mathbf{x}; {\cal L}_t)\;.  \label{eq: EGPmultiacq}
\end{align}
Upon querying the oracle for the label $y_{t+1}$ of instance ${\bf x}_{t+1}$, the labeled and unlabeled sets are updated, thus completing one iteration of the novel ``EGP-MultiAFs" approach, that is implemented as listed in Alg. 2.

\begin{figure*}[t]
\centering
	\includegraphics[width=0.988\linewidth]{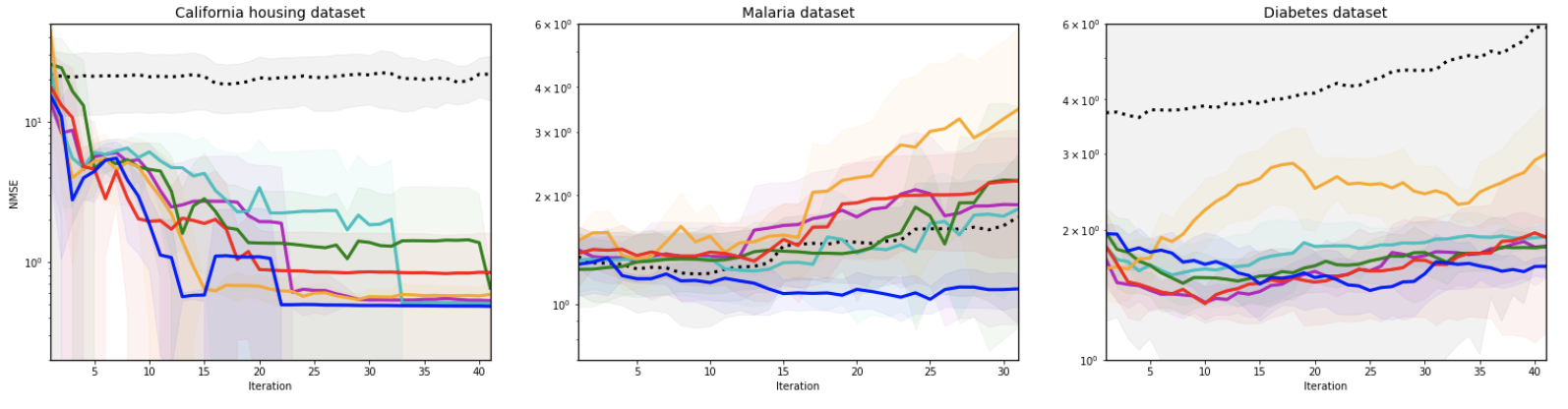}
	\includegraphics[width=0.99\linewidth]{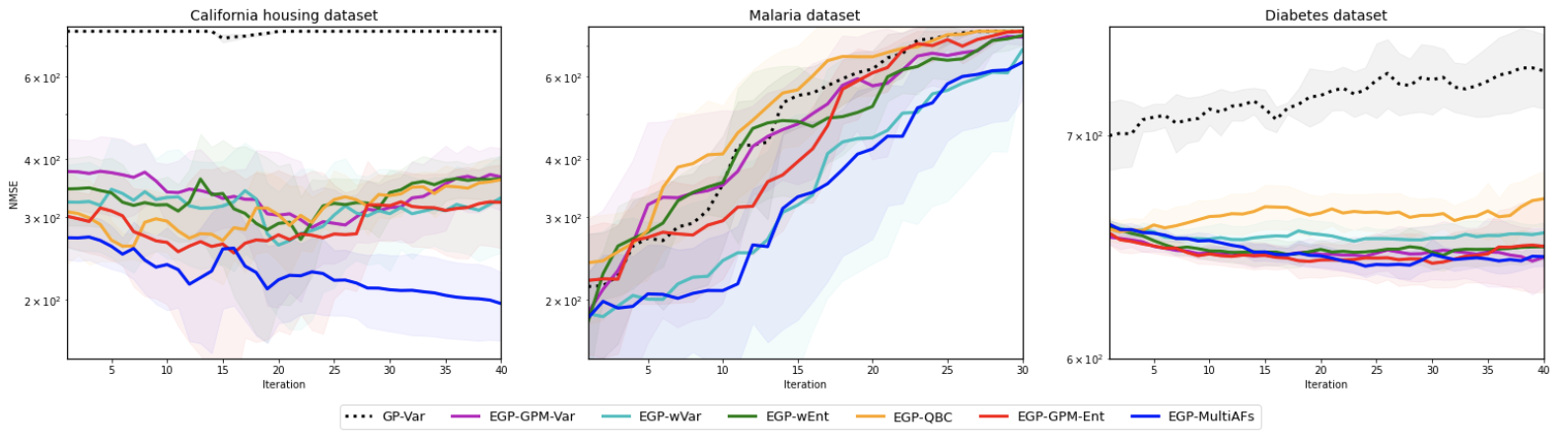}
		\vspace*{-0.22cm}
	\caption{NMSE (top) and NPLL (bottom) versus iterations for California housing, Malaria, and Diabetes datasets. } \label{fig:NMSE_real}
\end{figure*}


\section{Numerical tests}
In this section, the performance of the proposed EGP-based AL models will be compared against several benchmark synthetic functions, and it will be tested with real datasets ranging from biomedical to robotic based ones. Based on the novel EGP model, the innovative acquisition criteria to be tested are the ones described in \eqref{eq:w_var} - \eqref{eq:mix.ent.} and \eqref{eq: EGPmultiacq}, which henceforth will be abbreviated as ``EGP-wVar,'' ``EGP-wEnt,'' ``EGP-QBC,'' ``EGP-GPM-Var,'' ``EGP-GPM-Ent,'' and ``EGP-multiAFs,'' respectively. The competing baseline is a single GP model that utilizes the maximum variance (entropy) AF in \eqref{eq: singleGPvaracq} that has been extensively used in AL; see e.g., \cite{kapoor2007, pasolli2011, Schreiter2015SafeEF}. For all approaches, the few initially labeled data collected in $\mathcal{L}_0$ are utilized to obtain the kernel hyperparameters per GP expert by maximizing the marginal likelihood using the \textit{sklearn} package. The RF-based GP approximate models rely on $D=50$ RFs. For EGP-based approaches, the kernel dictionary $\mathcal{K}$ consists of radial basis functions (RBFs) with lengthscales $\{10^c\}_{c=-4}^6$. For the EGP-multiAFs approach, each $\alpha^k (\mathbf{x}; {\cal L}_t)$ in \eqref{eq: EGPmultiacq} is divided by its maximum value so that to range between 0 and 1.  

The performance of the competing methods is evaluated on a held-out test set $\mathcal{T}^e:= (\mathbf{x}_\tau^{e}, y_\tau^{e})_{\tau=1}^{T^e}$ (superscript $^e$ stands for evaluation) using two metrics. The first performance metric is the normalized mean-square error (NMSE) that for iteration $t$, is given by  
\begin{align*}
    \text{NMSE}_t := \frac{1}{T^{e}} \sum_{\tau = 1}^{T^e} (\hat{y}_{\tau|t}^{e} - y_\tau^{e})^2/\sigma_y^2 
\end{align*}
where $\hat{y}_{\tau|t}^e$ denotes the point prediction of test instance $\tau$, and $\sigma_y^2 := \mathbb{E} \| \mathbf{y}_T^e - \mathbb{E}\{\mathbf{y}_T^e\}\|^2$, where $ \mathbf{y}_T^e := [y_1^{e} \ldots y_T^{e}]^\top$. A second metric used to assess the associated uncertainty is the negative predictive log-likelihood (NPLL) expressed as 
\begin{align*}
    \text{NPLL}_t := - \log \; p(\mathbf{y}_T^e| \mathcal{L}_t, \mathbf{X}_{T^e}) 
\end{align*}
where the matrix $\mathbf{X}_{T^e} := [\mathbf{x}_1^e \dots \mathbf{x}_T^e]^\top$ collects the feature vectors of all $T^e$ test instances.  
All methods are tested over 10 realizations, and their average performance is reported along with the corresponding standard deviation. More details about the experimental set up can be found in Table 1 in the supplementary file. 

\begin{figure*}[!t]
    \centering
    \includegraphics[width=0.85\linewidth]{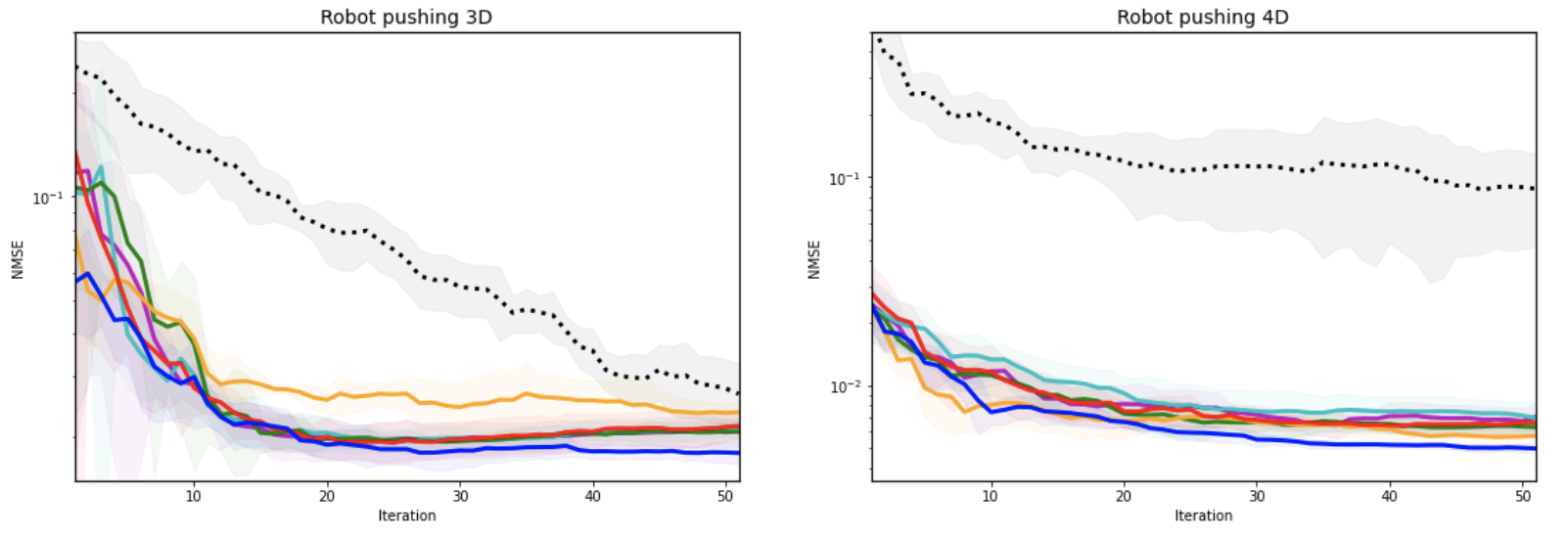}
    \includegraphics[width=0.85\linewidth]{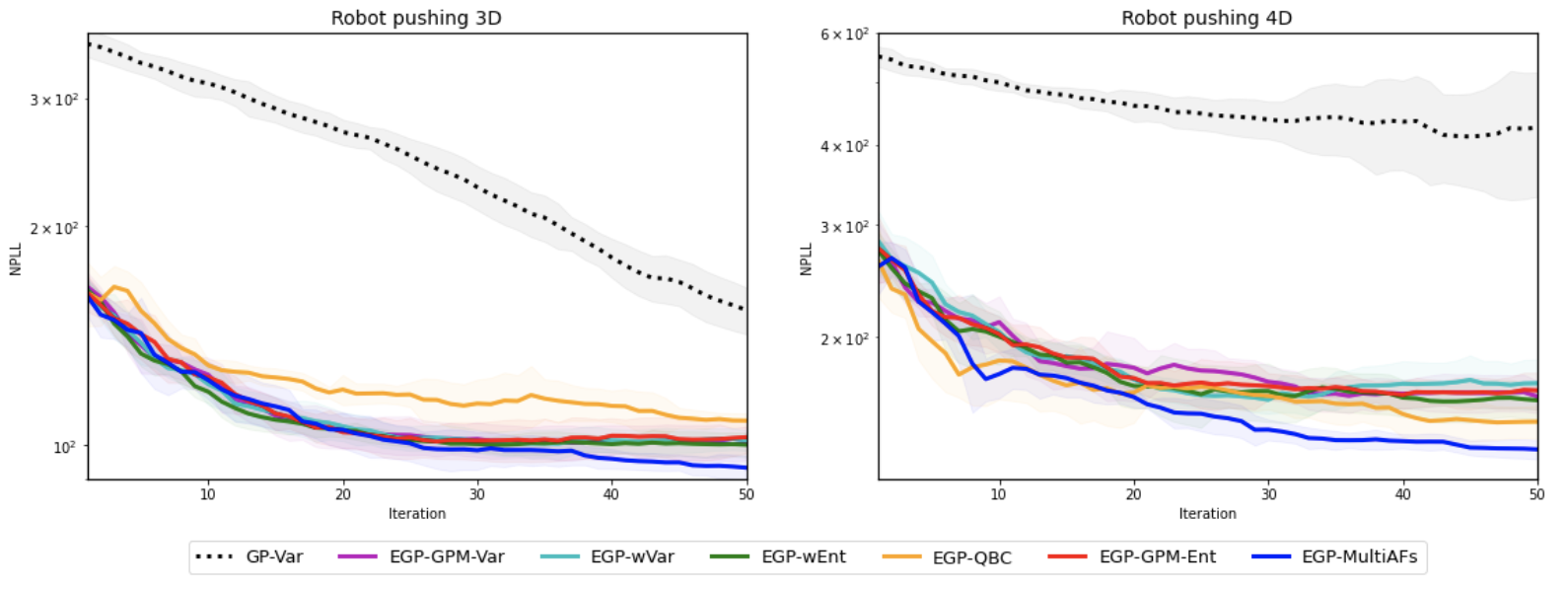}
    \vspace*{-0.3cm}
    \caption{NMSE (top) and NPLL (bottom) versus iterations for Robot Pushing 3D and 4D tasks.  }
    \label{fig:robot}
\end{figure*}

\subsection{Synthetic functions}
The tests here are run for known synthetic functions, including Ackley5D, Currin exponential, Branin, Gramacy and Higdon; see Table 2 in the supplementary file for their analytic expression. Fig. \ref{fig:NMSE_synFunc} demonstrates that all EGP-based approaches with a single AF outperform the GP-Var baseline in the Currin exponential and Gramacy functions, in terms of NMSE and most have superior performance in the remaining three datasets. In addition, all EGP-based methods enjoy the lowest NPLL in four out of five datasets, which corroborates the merits of having an ensemble of GPs and using them  in the corresponding acquisition criteria. Further exploiting an ensemble of AFs in the adaptive EGP-multiAFs approach, significantly improves the prediction performance, and also effectively quantifies the prediction uncertainty, thus rendering it the best performing approach over all datasets. 

\subsection{Real datasets}
All approaches here are tested on \textbf{California housing}, \textbf{Malaria}, and \textbf{Diabetes} real datasets. The last two deal with real medical data that are well motivated for AL because of the scarcity of labeled instances emanating from medical confidentiality. These datasets are described next.

\textbf{California housing dataset.} This set considers 8 features of districts in California, including demographic and location data, but also more general features such as average number of rooms and bedrooms per household, based on which a regression task is formed where the target variable is the median house price in these districts \cite{pace1997sparse}.

\textbf{Malaria dataset.} In this dataset, each instance corresponds to a specific location where the 8-dimensional feature vector consists of the latitude, longitude and some bioclimatic characteristics of this location, while the target variable is the infection rate of Plasmodium falciparum, the parasite that causes malaria \cite{weiss2019mapping,maddox2021conditioning}. 

\textbf{Diabetes dataset.} This dataset considers 10 characteristics of diabetes patients, including age, sex, body mass index, average blood pressure, and six blood-related measurements. The target variable is a metric that quantifies the disease progression in a single year \cite{efron2004least}.



It is evident in Fig.~\ref{fig:NMSE_real} that all EGP-based approaches markedly outperform the GP-Var baseline in terms of NMSE and NPLL in the California housing and Diabetes datasets, showcasing the significance of adopting the EGP model to estimate the learning function, along with the corresponding AFs. It is worth mentioning that although most EGP-based approaches are comparable in terms of prediction error in the California housing dataset, EGP-multiAFs outperforms all other methods in terms of NPLL.
Although the single GP approach is more effective in the Malaria dataset compared to the EGP-based ones with a single AF, EGP-multiAFs that judiciously combines the benefits of all AFs, enjoys the lowest NMSE. This illustrates the significance of properly adjusting AF weights in an online adaptive fashion. Regarding the NPLL metric, although it increases per iteration in the Malaria dataset, five out of six EGP-based approaches achieve lower NPLL compared to the GP-Var baseline.

\subsection{Robotic tasks}
The last experiments focus on a practical robotic task, where a robot pushes an object to a specific location~\cite{wang2017max}. Specifically, given as input the robot location $(r_\tau^x,r_\tau^y)$ and pushing duration $t_\tau^p$ at slot $\tau$, the object ends up in a location $\mathbf{o}_\tau := (o_\tau^x,o_\tau^y)$. We form a regression task where the goal per slot $\tau$ is to map the $3\times1$ feature vector $\mathbf{x}_\tau := [r_\tau^x, r_\tau^y, t_\tau^p]^\top$ to the target variable $y_\tau := ||\mathbf{o}_\tau - \mathbf{d}||_2$, with $\mathbf{d}:= [d^x,d^y]$ denoting a pre-defined constant position vector, yielding the \textbf{Robot pushing 3D} dataset. This is of practical interest in various robotic problems such as obstacle avoidance, where $y_\tau$ is desired to be greater than a pre-defined threshold $y^{\text{th}}$. Augmenting the feature vector $\mathbf{x}_\tau$ with the additional feature $r_\tau^\theta$ that represents the pushing angle, yields the \textbf{Robot pushing 4D} dataset.  

Fig.~\ref{fig:robot} depicts the NMSE and NPLL at each iteration of all competing AL approaches for the Robot pushing 3D and 4D tasks respectively. It is evident that all EGP-based approaches enjoy remarkably lower NMSE and NPLL compared to the single GP based AL counterpart in both datasets, with the EGP-MultiAFs consistently being the best-performing one. This implies that in these practical robotic tasks, the function expressiveness offered by the advocated EGP model and the ensuing innovative acquisition criteria considerably improve the prediction performance providing also quantifiable prediction uncertainty.     
\section{Conclusions}
\vspace*{-0.2cm}
This work advocated a weighted ensemble of GPs as the statistical model in AL. By adapting the weights of individual GPs, the EGP model selects the appropriate kernel on-the-fly as new labeled data are included incrementally. Building on the novel EGP model, several AFs have been devised based on different criteria. Combining the candidate EGP-based AFs with weights being adjusted in an adaptive manner, the AL performance can be further robustified. Tests on synthetic functions and real datasets showcase the merits of weighted ensembles of GPs and AFs in AL.

\bibliographystyle{abbrv}
\bibliography{EGPAL_arxiv-main}

\clearpage
\onecolumn
\begin{center}
  \Large{\textbf{Supplementary file}}
\end{center}

\begin{table*}[!h]
	\caption{Additional experimental details} \label{table:exper_details}
	\begin{center}
\begin{tabular}{c|c|c|c|c|c}
			\hline
			\hline
	Dataset &  $\mathcal{L}_0$ size & $\mathcal{V}$ size & $\mathcal{U}_0$ size & $\mathcal{T}$ size & $\eta$   \\
			\hline
			\hline 
Ackley-5D 	 & 10	 & 50  & 500  & 100 &  1   \\
			\hline 		
Branin 		 & 10	 & 50   & 500   & 100    &  100   	   \\
			\hline
Currin exponential 	 & 10 	 & 50  & 500   & 100   & 100    \\
	        \hline
Gramacy 	 & 10	 & 50  & 500  & 100   &   100     \\   
	        \hline
Higdon 	 & 10	 & 50  & 500   & 100   & 100      \\    
	        \hline
Diabetes 	 & 15	 & 55  & 261  & 111   &  100     \\    
	        \hline
Malaria	 & 40	 & 60  & 500  & 680   & 20      \\    
	        \hline
Robot pushing 3D 	 & 20	 & 50  & 500  & 200   &  20     \\    
	        \hline
Robot pushing 4D	 & 20	 & 50  & 500  & 200   & 20       \\    
	        \hline
California housing 	 & 50	 & 70  & 1000  & 1032   &  0.05     \\    
	        \hline
		\end{tabular}
	\end{center}
\end{table*}

\begin{table*}[!h]
	\caption{Analytical expression of all synthetic functions} \label{table:Analytic_exp}
	\begin{center}	\begin{tabular}{ l p{.78\linewidth} p{.78\linewidth} }
			\hline
			\hline
		Function & 	Analytical expression  \\
			\hline
			\hline 
Ackley-5D 	 & $ -20 e^{-0.2\sqrt{(x_1^2+x_2^2+x_3^2+x_4^2+x_5^2)/5}} -e^{(\text{cos}(2\pi x_1)+\text{cos}(2\pi x_2)+\text{cos}(2\pi x_3)+\text{cos}(2\pi x_4)+\text{cos}(2\pi x_5))/5} + 20 + e^{1}$	  \\
			\hline 		
Branin 	&  $(x_2 - 5.1 / (4 \pi^ 2) x_1^ 2 + 5 x_1 / \pi - 6) ^ 2 + 10 (1 - 1 / (8 \pi))\text{cos}(x_1) + 10$	   \\
			\hline
Currin exponential & $(1 - e^{-1/(2x_2)})(2300x_1^3 + 1900x_1^2 + 2092x_1 + 60) / (100x_1^3 + 500x_1^2 + 4x_1 + 20)$\\
	        \hline
Gramacy &  $\text{sin}(10\pi x)/(2x) + (x-1)^4$	     \\   
	        \hline
Higdon & 	$\text{sin}(2\pi x/10) + 0.2\text{sin}(2\pi x/2.5)$   \\    
	        \hline
		\end{tabular}
	\end{center}
\end{table*}

\end{document}